\documentclass{article}
\input{config.sty}
\input{abbr.sty}

\usepackage{palatino}
\usepackage{subcaption}
\usepackage{anyfontsize}
\title{\fontsize{15pt}{15pt}\selectfont\textbf{To think inside the box, or to think out of the box?\\
Scientific discovery via the reciprocation of insights and concepts}\par}

\author{
Yu-Zhe Shi$^{1,2\,\star{}\,\textrm{\Letter}}$, Manjie Xu$^{1,2\,\star{}}$, Wenjuan Han$^{2}$, Yixin Zhu$^{1\,\textrm{\Letter}}$
\vspace{0.5em}\\
\small$^1$Institute for AI, Peking University\quad{}
\small$^2$PersLEARN\\
\small$^\star{}$Equal contributors\quad
\small$\textrm{\Letter}$\phantom\,\,\texttt{syz@perslearn.com, yixin.zhu@pku.edu.cn}
}
\date{}

\usepackage{indentfirst}
\usepackage{multirow}
\usepackage{colortbl}
\usepackage{hhline}

\begin{document}
\maketitle

\setstretch{0.995}

\begin{abstract}
    \noindent If scientific discovery is one of the main driving forces of human progress, insight is the fuel for the engine, which has long attracted behavior-level research to understand and model its underlying cognitive process. However, current tasks that abstract scientific discovery mostly focus on the emergence of insight, ignoring the special role played by domain knowledge. In this concept paper, we view scientific discovery as an interplay between \emph{thinking out of the box} that actively seeks insightful solutions and \emph{thinking inside the box} that generalizes on conceptual domain knowledge to keep correct. Accordingly, we propose Mindle, a semantic searching game that triggers scientific-discovery-like thinking spontaneously, as infrastructure for exploring scientific discovery on a large scale. On this basis, the meta-strategies for insights and the usage of concepts can be investigated reciprocally. In the pilot studies, several interesting observations inspire elaborated hypotheses on meta-strategies, context, and individual diversity for further investigations.
\end{abstract}

\section*{Introduction}
How do scientists come up with novel ideas that lead to significant discoveries? Psychologists have been working long to understand the underlying cognitive processes \citep{sep-scientific-discovery} to facilitate the progress of scientific discovery \citep{campbell1960blind}. Among the diverse philosophical theories interpreting \emph{discovery}, the one mostly being cited is that discovery refers to \emph{the eureka moment}, \aka \emph{the Aha! moment}, of having a new \emph{insight} \citep{auble1979effort,kounios2009aha}. Originating from problem-solving, insight is the process that reconstructs the representation of the target problem. Given the insight, the solution can be achieved much more straightforwardly than that before the reconstruction has been done \citep{ohlsson1984restructuring}. People tend to follow prior knowledge when solving a problem because experience shows this may lead to success \citep{ollinger2008investigating}. But after times of trial-and-error, people can predict the error of current problem representation \citep{dubey2021aha}, and this may be the eve of a sudden coming of the Aha! moment. Studies have shown that solutions discovered by insights usually be more promising than those generated by analytical approaches \citep{salvi2016insight}, though the latter requires much more workload than the former---this echoes how specifically adapted representations outperform prior ones on novel problems. 

Meanwhile, the prior representation is also necessary, for it provides the relevant domain knowledge for the problem---insight is just useless without the sense of how to deal with the problem. This contrast becomes crucial in terms of scientific discovery---given a scientific inquiry as the target problem, the entry is the purported paradigm of the domain that seems relevant to the problem, which is deeply rooted in \emph{domain knowledge} (\ie, atoms, theories, and claims) and shapes the meta-cognitive strategic knowledge (\ie, methodologies). Hence, representation reconstruction is extremely hard when the target problem becomes a scientific inquiry because that at least means generalization from a scientific to another given few observations \citep{tenenbaum2001generalization}, and even means a paradigm shift \citep{kuhn1970structure}; but without relying on domain knowledge, a scientist can go toward nowhere because all ideas come from somewhere.

\begin{figure}[t!]
    \centering
    \includegraphics[width=\linewidth]{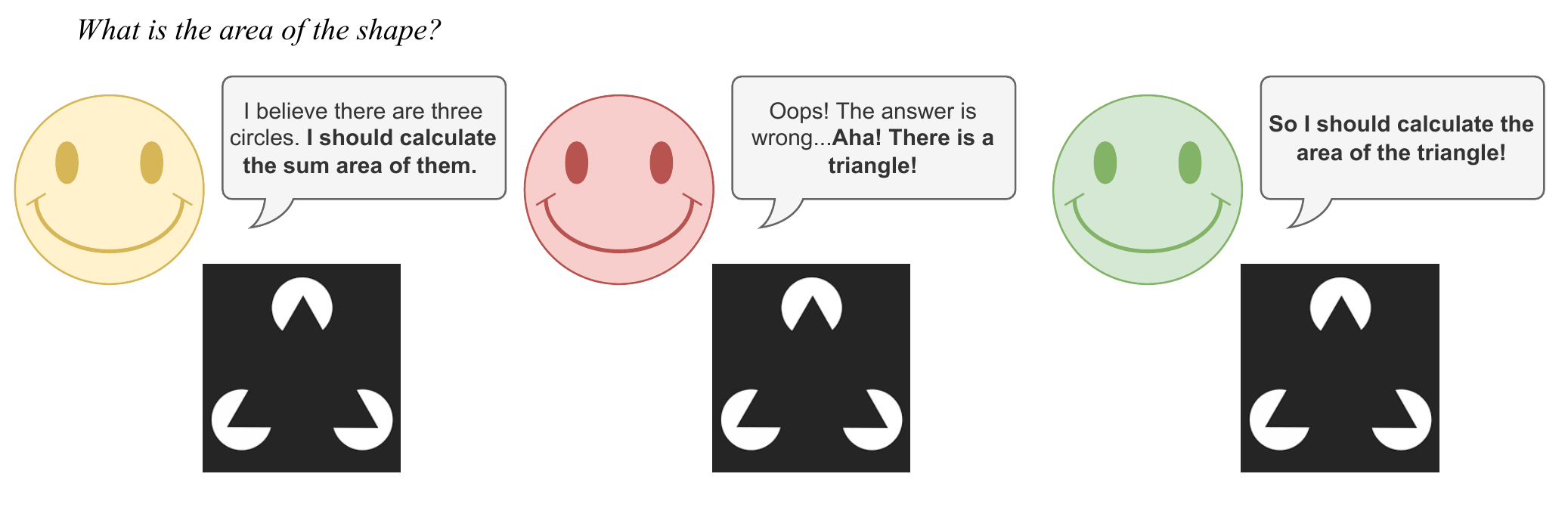}
    \caption{\textbf{Overview of insight in scientific discovery.} In a classic Gestalt problem, a problem solver first uses domain knowledge to analyze the problem, then seeks for insight once she gets trapped; after reconstructing the problem representation, she again uses domain knowledge to reach the solution. In this case, though domain knowledge constrains the thinking, it serves as the vehicle toward the target.}\label{fig:overview}
\end{figure}

The dilemma we are facing is ubiquitous in scientific discovery. On the one side, we must \emph{think out of the box}, such that to avoid missing the flashed-by insights; but unconventional ways of thinking may also lead to ridiculous solutions, taking a detour even compared with analytical solutions. On the other side, we have to \emph{think inside the box} because domain knowledge keeps us aware of what we are doing and where we are going; but following an established paradigm totally restricts our mind in the prior representation of the problem (see \cref{fig:overview}). To achieve solutions successfully and efficiently, the two mindsets should interplay with each other. When and where should this happen?

Many scientists deal with such an interplay well---they gain insights from the eureka moment, which are later developed into representative scientific discoveries, such as the development of Einstein's special relativity \citep{einstein1982created}, the discovery of Kekule structure \citep{gruber1981relation}. This pattern is also found in many works throughout the life of Gauss \citep{dunnington2004carl}. Both historical experiences and experimental results drive the interest in understanding how insight is obtained and making computational models \citep{langley1988computational}, to get close to the ultimate goal---automated production of insights that improve scientific discovery. Unfortunately, stories of scientists cannot lead to concrete modeling and evaluation work at the behavior level rather than the metaphysical level, for post-hoc simulating how scientists disentangle meta-cognitive strategies from domain knowledge is difficult and imprecise. Hence, we are on the request of an experimental environment that abstractly simulates the process of scientific discovery---the domain knowledge should be crucial for solving the problem and should be general enough to carry out large-scale behavioral studies \citep{almaatouq2021scaling}, without losing of group convergence or individual diversity. To the best of our knowledge, we are the first to explicitly consider the interplay between \emph{insight-seeking} and \emph{domain-knowledge-relying}. Hence, we propose \textbf{Mindle}\footnote{Visit \url{mindle.cn} to interact with the web-based user interface.}, a semantic searching game that triggers scientific-discovery-like thinking spontaneously, as infrastructure for exploring scientific discovery on a large scale, filling the gap in the literature.

\section*{The reciprocation of insights and concepts}

Based on the dilemma over insight-seeking and domain-knowledge-relying, the most critical feature that distinguishes scientific discovery from normal insight problem-solving is that domain knowledge plays a crucial role in both empowering the solution to be correct and restricting the emergence of insight solutions efficiently. Hence, to understand scientific discovery \emph{inside the box}, we should understand how the organization of concepts in domain knowledge affects meta-cognitive strategies in advance. Conversely, to unveil the process of scientific discovery \emph{outside the box}, we should look into how insightful decisions intervene in using concepts. This bidirectional pathway echoes the reciprocation of insights and concepts by identifying these two questions: (1) How a problem grounded on conceptual knowledge improve the study of insight problem-solving? (2) How does the usage of conceptual knowledge driven by insight problem-solving improve the study of knowledge representation? Below, we sketch Mindle by answering these questions. 

\paragraph{Concepts improve the study of insight problem solving}

Relying on conceptual knowledge is not an obstacle to investigating scientific discovery, but a better chance for understanding insight problem-solving. Current insight problem-solving tasks have been long troubled by the subjects' unawareness of the meta-cognitive strategies they have used \citep{metcalfe1987intuition}. Though this inability naturally unveils the sudden come of insights, it is avoidable by improving experimental tasks. Some current insight problems focus on stimulating the eureka moment, such as the nine-dot problem \citep{kershaw2004multiple}, the matchstick arithmetic \citep{knoblich1999constraint}, and the eight-coin problem \citep{ormerod2002dynamics}---these tasks provide highly confined problem space such that subjects can solve the problems without applying any semantics or commonsense knowledge other than the specific background knowledge given by the problem settings. Although such designs are motivated to disentangle representation reconstruction from solution cleanly, it makes interpreting the solutions from the trajectories hard, since the trajectories can only be mapped to the given problem settings. Hence, if we expect mapping behavioral trajectories to meta-strategic knowledge, a generally understandable semantics of problem context is necessary, such as conceptual knowledge in human language. Other insight problems introduce general semantics, including insight physical problem solving (\ie, intuitive physics as semantics) \citep{allen2020rapid} and remote association test (\ie, word association as semantics) \citep{mednick1963research}. However, these tasks come in a one-to-one input-output fashion, where the measured behavior is directly generated from the single stimuli one-step, thus hard to track the representation change. This is crucial in scientific discovery because there are usually multiple steps of insights that lead to the target \citep{moszkowski1972conversations}. In contrast to one-shot problem solving, scientific discovery is more like a path-finding process where the navigation map changes every time reaching a critical point. Putting the two reasons together, Mindle should equip with general-understandable semantics in the context of the target problem. 

\paragraph{Insight problem-solving improves the study of conceptual knowledge} 

The domain knowledge of sciences is believed to be organized as conceptual knowledge \citep{hiebert1986conceptual,rittle2001developing}. Conceptual knowledge systematically combines declarative and procedural knowledge, consisting of both facts about the concepts and active processes about how concepts interact with each other \citep{abend2008meaning}. Though there are many perspectives on concept representation, we take the theory theory as a prerequisite because it is the most accepted theory in terms of scientific knowledge representation \citep{gopnik1994theory}, thus we can mimic the domain knowledge under the form of theory theory. One implementation of theory theory is that concepts are maintained in a fully-connected network, where each concept is related to all other concepts in the set---many calculi on fully-connected graphs, such as the general pattern theory \citep{grenander2012calculus}, can be applied to formalize the operations over concepts. Such tools help describe scientific knowledge and meta-strategies in a computable way. Thanks to that, the theory theory can also model how a child acquires concepts in cognitive development \citep{carey1985conceptual,gopnik1997words,carey2009origin}; people similarly organize normal concepts to organize scientific knowledge. Hence, we may use such fully-connected networks statistically extracted from natural language corpus to simulate the domain knowledge of sciences, so we can carry out large-scale behavioral studies. Most interestingly, there is a major feature shared by both normal and scientific knowledge---though the semantics of concepts and relations are static and invariable viewing knowledge in the world holistically, they may be highly overloaded according to diverse context, task utility, and inner preference, from the view of individuals \citep{wang2021idiosyncratic}. In this way, the compounded concepts are projected to simplified semantic attribute spaces \citep{grand2022semantic}, which are much more tractable to be processed due to the rational use of limited cognitive resources \citep{gershman2015computational,lieder2020resource,ho2022people}. On this basis, conceptual knowledge in scientific discovery should be studied in a dynamic way rather than a static way. However, most current behavior-level experimental methods on semantic understanding of concepts are confined to general and fixed contexts \citep{huth2016natural,wang2020two}, in a straightforward way that is given the stimuli (input words) to obtain the descriptions or similarity judgments (output measurements) directly; other studies on the use of concepts, such as memory replay and human reinforcement learning \citep{momennejad2017successor}, mostly focus on the short-term memory for skill learning without retrieving from long-term memory that has been already formed. To capture the two features, Mindle should test the representation of conceptual knowledge in sequential decision-making.

\vspace{\baselineskip}
In summary, we profile Mindle with two unique features to simulate the process of scientific discovery (see \cref{tab:analogy} for details), and importantly, the two work reciprocatively: (1) Mindle should equip with a general-understandable conceptual knowledge as domain knowledge to help interpret the process of insight problem solving; (2) Mindle should equip with a sequential decision-making task to stimulate the flexible dynamic use of conceptual knowledge. 

\begin{table}[t]
    \centering
    \caption{\textbf{The analogy of scientific discovery to solving Mindle}}\label{tab:analogy}
    \resizebox{\linewidth}{!}{%
        \begin{tabular}{ccc}
        \toprule
                                 & \textbf{scientific discovery}                                    & \textbf{solving Mindle }                                    \\
        \midrule
        \textbf{target}                   & solve a scientific query                                & find out the secret word                           \\
        \textbf{output}                   & the hidden answer is shaped by \emph{the known}                & the secret word is among \emph{the known}                 \\
        \textbf{problem abstraction}   & path searching from status quo to \emph{the unknown}           & path searching from current guess to \emph{the unknown}   \\
        \textbf{problem context}          & conceptual scientific domain knowledge                  & conceptual knowledge from natural corpus           \\
        \textbf{knowledge representation} & concepts connected under logical or intuitive relations & concepts connected under intuitive relations       \\
        \textbf{maintained representation}           & generalizing from one scientific concept to another       & generalizing from one concept to another \\
        \textbf{reconstruction}           & changing the domain knowledge or methodology used       & changing meta-strategy in action or semantics level \\
        \textbf{rationality}              & should be studied from specific perspectives            & should be used in specific semantics subspaces     \\
        \textbf{diversity}                & scientists with different background think differently  & people with different background think differently \\
        \textbf{accessibility}            & captured by only a few individuals                      & captured by most individuals                       \\ \bottomrule
        \end{tabular}%
    }%
\end{table}

\section*{Mindle: an infrastructure for large-scale studies on scientific discovery}

Given the specific features that Mindle should capture, we describe how to implement Mindle. As the infrastructure for large-scale behavioral studies, Mindle should meet three elementary satisfaction: (1) providing appropriate tasks that abstract the target real problem to be studied; (2) providing correct computational models to evaluate the results; (3) the abstract task itself should be natural rather than artificial, interesting rather than boring, to make sure that subjects are easy to get into the task; (4) the task is easy to be propagated and is robust to unexpected user behaviors. Since the appropriateness of task abstraction has been illustrated in detail, we describe how Mindle meets the last three.

\paragraph{The semantic searching game}

Mindle requires the participants to dig out a hidden secret word. In every single challenge, the participant is given only a starting word. In each guess, the participant inputs a guessed word, and Mindle outputs a score (the given starting word can be viewed as the initial guess). The score indicates how \emph{far} is the current guess to the target secret word, on a 0 to 100 scale---when the score is more close to 100, the guess is more close to the target. Once the secret word is hit, one challenge ends. The participant can choose to quit at any time during the challenge. The guesses can be entered by the participant through an input text bar or be selected by the participant through some given options; the policy for proposing options will be described later. A variant of challenge mode comes with a hint implying what topic the secret word is related to. The topic can either be abstract or concrete, such as \emph{kitchen supplies}, \emph{classical music}, or \emph{freedom}, which confines the problem space.

\paragraph{Score and vocabulary}

The score is provided by the cosine similarity between the embedding vectors of the current guess and the target word. The embedding vectors can be generated by arbitrary embedding methods, such as Word2Vec \citep{mikolov2013distributed}, Skip-gram \citep{mikolov2013efficient}, or Glove \citep{pennington2014glove}. The score transforms the 0 to 1 scale of cosine similarity to a 0 to 100 scale. The vocabulary we used here is a subset consisting of about 40K most frequently used concepts in the natural corpus. Here we denote the set as $C\in\mathcal{C}$ where $\mathcal{C}$ is the space of all concepts. In bar-entering mode, participants would be informed if the input guess is out of $C$.

\paragraph{Conceptual knowledge representation}

We have mentioned that conceptual knowledge can be represented as a fully-connected network. It can be implemented as a directed graph encoding an adjacent matrix. The graph is obtained from natural corpus through a Transformer model \citep{vaswani2017attention}. The connection weights in the graph capture the co-occurrence frequency for each pair of concepts. Let graph $G=\langle C, C\times C \rangle$ be the system of concepts, each node $c_i\in C$ denotes a concept, $c_k,k\neq i$ is a possible related concepts that shapes $c_i$, and $w(c_i,c_k),k\neq i$ indicates the weight $c_i$ is shaped by $c_k$ among all related concepts. The higher $w(c_i,c_k)$ is, the semantics of $c_i$ is more influenced by that of $c_k$. Note that usually $w(c_i,c_k)\neq w(c_k,c_i)$ because the weights indicates the conditional probability $p(c_i|c_k)=w(c_i,c_k) / \sum_{j\neq i} w(c_i,c_j)$. Intuitively, this can be viewed as the probability of describing the concept $c_i$ by concept $c_k$ in a rational fashion \citep{frank2012predicting}. $p(c_i)$ is a vector that has $p(c_i|c_k)$ as its $k$-th dimension. And since the statistical feature is obtained from the general natural corpus, the real conceptual knowledge distribution in individuals' minds is more or less different because that is highly conditioned on diverse individual prior experiences. In this way we define all general and individual operations over concepts as formal operations on a graph.

\paragraph{Behavior modeling}

Solving Mindle can be ideally modeled as a \acf{mdp}. An \acs{mdp} is a tuple $T=(S,A,P,R,\rho)$, where $S=C$ is the set of states, $A=C$ is the set of actions, $P:S\times A\times S\mapsto [0,1]$ is the transition probability function $s_{t+1}\sim P(\cdot|s_t,a_t)$, $R:S\times A\times S\mapsto\mathbb{R}$ is the reward function (the game score here plays the role of reward) indicating in what condition the task is solved, and $\rho$ is the initial guess $\rho$ where $s_0\sim\rho$. Under this formulation, solving Mindle is finding the policy $\pi=(s_0,a_0,s_1,a_1,\dots)$ that generates trajectory $\tau\sim\pi$ optimizing the objective $\max_{\pi}\mathbb{E}_{\tau\sim \pi}[\sum_{t=0}^\infty\gamma^t R(s_t,a_t,s_{t+1})]$. But this modeling is extremely ideal because the spaces of state and action are much smaller than the whole set of concepts, due to highly limited memory and attention slots. Hence, a mask $M$ can be applied to both $S$ and $A$ that $S\cdot M\subset C$ and $A\cdot M\subset C$. $M$ can either be predefined by heuristics (will be introduced next) or be extracted from trajectories generated by subjects in pre-experiments. 

\paragraph{Similarity, rule, and the unrelated}

\begin{wrapfigure}[18]{R}{0.38\linewidth}
    \centering
    \vspace{-5pt}
    \includegraphics[width=\linewidth]{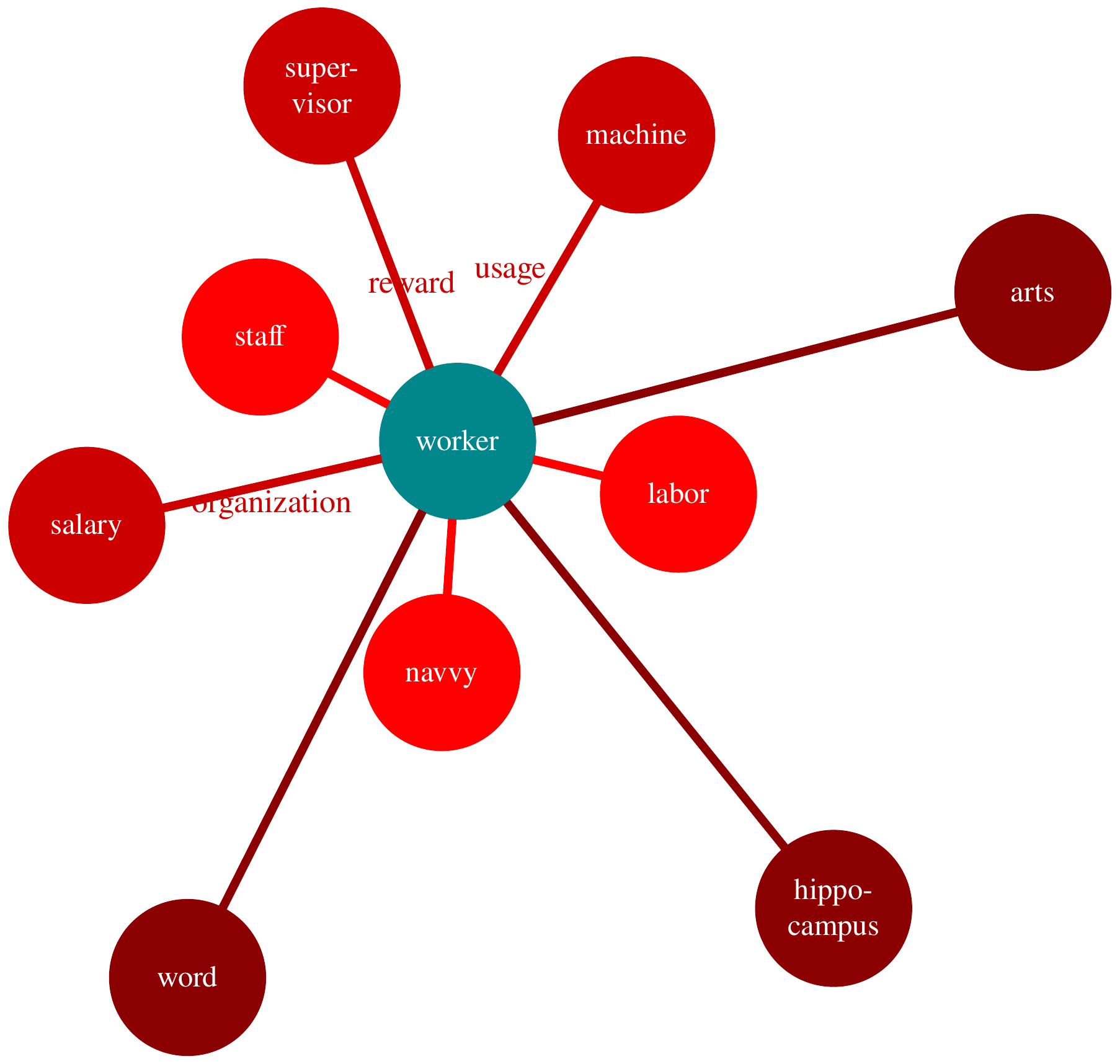}
    \caption{\textbf{An example of conceptual knowledge representation.}}
    \label{fig:example}
\end{wrapfigure}

Given the current guess, where can we go? Take \emph{worker} as the exemplar current guess, there are three types of candidate actions (see \cref{fig:example}): (1) Concepts that are highly similar to current guess, \eg, \emph{labor}, \emph{navvy}, and \emph{staff}. (2) Concepts that are highly related to current guess by specific rules, \eg, \emph{machine} (by rule \emph{usage}), \emph{salary} (by rule \emph{reward}), and \emph{supervisor} (by rule \emph{organization}). (3) Concepts that are hardly related to current guess by any mean, \eg, \emph{arts}, \emph{word}, and \emph{hippocampus}. Suppose we have K proposals in each type and denote $c_t$ and $c_{t+1}$ as current and next guess, respectively. The similar concepts are generated by top-K $\max_k cos(\text{Vec}(c_t),\text{Vec}(c_k)),k\neq t$, the related concepts are generated by top-K $\max_k w(c_t,c_k),k\neq t$, and the unrelated concepts are generated by top-K $\min_k w(c_t,c_k),k\neq t$. Most similar concepts are synonyms to $c_t$, which are not highly related to $c_t$; exceptions exist, thus making a filter afterward. When selecting related concepts, we try to maximize the diversity of the candidate concepts by filtering out duplicate concepts generated by the same rule, \eg, \emph{boss} and \emph{supervisor} (by rule \emph{organization}). We implement this by applying Agglomerative clustering to the candidate concepts and selecting from the root \citep{murtagh2014ward}. The three types are actions in the higher level of $A$, which can be observed directly and easily identified to decision patterns, such as \emph{persistent searching in a local minima} or \emph{flexibly jumping between locals}. Besides analyzing behavior data, the graph pruned by the three modes is also used to control the hardness of challenges by tuning the length of the shortest path and the number of possible paths traversing from the initial guess to the target word.

\paragraph{Evaluation metrics}

One significant indicator to be evaluated is the emergence of the eureka moment. We use both absolute and relevant measurements. First, we identify the Aha! moments in a single trajectory $\tau=(s_0,a_0,a_1,s_1,\dots,a_{-1},s_{-1})$. First, we calculate the first-order reward difference between adjacent trials $\Delta r(t)=r_{t+1}-r_t$. For action $a_t$, if there exists $i<t$ that $\Delta r(i)\geq\Delta r(t)$, let $r(i:t)=\sum_{k=i+1}^t r_k / (t-i)$, if not then let $i=0$; if there exists $j>t$ that $\Delta r(j)\geq\Delta r(t)$, let $r(t:j)=\sum_{k=t+1}^j r_k / (j-t)$, if not then let $j=-1$. We have $\Delta a(t)=r(t:j)-r(i:t)$ as the cumulative difference between the critical points. Those critical points with high $a(t)$ tend to be eureka moments from the view of a single trajectory. Also, we observe the trajectories in a counterfactual fashion, and we ask what would happen if other actions have been taken. The action space can be either the concept space $C$ or masked $C\cdot M$ in the low level, or the space of the three high-level action types (similarity, rule, and unrelated). Then, we define the updating rate as $\min\{0,1-\min_{a\in A} R(s_t,a,s_{t+1})/R(s_t,a_t,s_{t+1})\}$. Thus, the current guess can be viewed as an intervention to the status quo following the semantics gradient, supporting investigations on semantics overload. The evaluation metrics can be flexibly modified to meet the need of specific experiments. 

\paragraph{Playability as a game} 

In the post-interview of our pilot study (25 subjects, 11 are female), after 3 challenges per subject, 24 in 25 subjects think that \emph{playing Mindle is interesting} and 18 in 24 subjects think that \emph{I want to play Mindle everyday}; 15 in 25 have succeeded at least one challenge and 8 in 15 have experienced at least one eureka moment in at least one challenge. All solved challenges are finished in about 80 steps of guessing on average, and the number of insightful solutions is about 50. These results imply that Mindle is attracting enough with an appropriate success rate, having the potential to propagate with ease. Also, Mindle is unique for its sense of infinite answer space and harmless trial \citep{hamari2007gamification}. Thus, Mindle can be played through a long term, even though a challenge is disrupted in the middle---instead, a long period of discontinuous thinking may even stimulate the Aha! moment. 

\section*{Case Studies}

In this section, we discuss three interesting observations from the pilot study. These phenomena may inspire elaborated hypotheses that lead to further investigations. This shows how Mindle empowers large-scale experiments.

\paragraph{Thinking out of the box: action level}

Some challenges succeed due to action-level insights. The trajectories have a common point---participants switch smoothly between \emph{searching in a local minima} or \emph{making traversals between locals globally}. Once they have been optimizing a local minima for a time without gaining a significant score increase, they changed to jump randomly between concepts that seem unrelated to each other, until they hit a local with a much higher score. Then they settle down again to optimize the guess locally with synonyms or similar concepts. The eureka moments usually come when hitting a \emph{hot solution} in the global random jump. Interestingly, the concepts in the trajectories are highly different from each other, indicating that such behavior pattern is not constrained by semantics, but be driven by the inner preference of participants. That is, no matter what concepts I am guessing, I just switch between local search and global jump flexibly. This lead to a hypothesis: Do people apply meta-strategies ignoring the problem context?

\paragraph{Thinking out of the box: semantic level}

Some challenges succeed due to semantic-level insights. This happens when participants hit a local with a relatively high score, which is easy to believe that target lies in this local. Affected by prior experience, participants tend to search in a subspace of the semantics space, where the selected concepts are projected onto a plate of reduced semantics space. For example, a participant has guessed \emph{school}, \emph{class}, and \emph{grade}, where she projects the concept to the semantics subspace of school-related concepts. However, these words cannot help go further. The participant decides to project the anchor concept, say \emph{class}, to another semantics space, say \emph{computer-related terminologies}. Then, she guessed \emph{type} and unexpectedly takes a large step toward the target. Hence, she understands that she has been trapped in a semantics subspace. This case shows that representation reconstruction can also change the semantics subspace. Hence, we come up with another hypothesis: Do people apply meta-strategies according to the problem context? This hypothesis seems to be in contrast to the one in the last paragraph, but combining these two together, we have a comparative hypothesis, which is more related to our big picture on scientific discovery: Do people use meta-strategy as policy regardless of context, or subject to the subjective understanding of context semantics?

\paragraph{Thinking inside the box: semantic level}

Some challenges succeed through analytical solutions, especially when the participant, fortunately, reaches the right track at the start. The participant optimizes a gradient of the semantics landscape in mind. The gradient can be extremely flexible---for example, hierarchy, \emph{arts} to \emph{painting} to \emph{gallery}; extent, \emph{large} to \emph{larger} to \emph{largest}; or the distance with human, \emph{human} to \emph{chimpanzee} to \emph{monkey}. The hypothesis space for such a gradient is almost infinite because the semantics space is in very high dimensionality \citep{grand2022semantic}. Since the choice of the gradient is subjected to personal cognitive bias, the trajectories for the same challenge can be highly diverse. This echoes the diverse \emph{mindsets} of different genres of science. And compared with previous work on testing personal diversity in knowledge representation \citep{wang2021idiosyncratic}, Mindle (1) stimulates the spontaneous use of the commonsense knowledge, in contrast to other experimental paradigms that probe human knowledge representation explicitly; (2) empowers the scaling-up of pilot studies, both in the broadness of semantics and the diversity of subjects, to obtain more elaborated results on the landscape that where people converge or diverge on concept representation. By recovering all trajectories generated by the same group of participants, we can build a computational model that captures their semantics landscape, \ie, a function that outputs the sense of \emph{which concepts are more similar or more related to each other than to others}. The function can be approximated through inverse reinforcement learning \citep{abbeel2004apprenticeship}, thus we can analyze group diversities of concept representation quantitatively. In this way, we may reverse-engineer the organization of conceptual knowledge in peoples' minds.

\paragraph{Combining the three hypotheses} 

Testing the three hypotheses helps us understand the interplay between insight-seeking and domain-knowledge-relying. First, on a confined problem space, we test the existence of these thinking patterns, by stimulating the spontaneous use of action-level metastrategies, semantic-level metastrategies, and semantic-level landscape optimization. After this, we study given different constrained topics, are people tend to emerge and converge to a set of similar parameters that controls the interplay. Assuming that insight-seeking and domain-knowledge-relying are on the two ends of a continuum, one possible hypothesis is that people rationally control the interplay according to their uncertainty on the use of domain knowledge---high uncertainty on \emph{knowing what} may lead to reconstruction at action level; high uncertainty on \emph{deciding which} may lead to reconstruction at semantic level; and low uncertainty may lead to maintaining the current representation. Such intuition defines the \emph{balance point} between the two ends, thus irrational behaviors, \eg, more close to insight-seeking, can be identified comparing with the rational case. On this basis, we scale up the behavioral studies to generalize the results to larger groups of individuals, and also collect a large dataset of trajectories and reverse-engineer the semantic landscapes in the open domain. 

\begin{figure}[t!]
    \centering
    \begin{subfigure}[t]{0.33\linewidth}
        \centering
        \includegraphics[width=\linewidth]{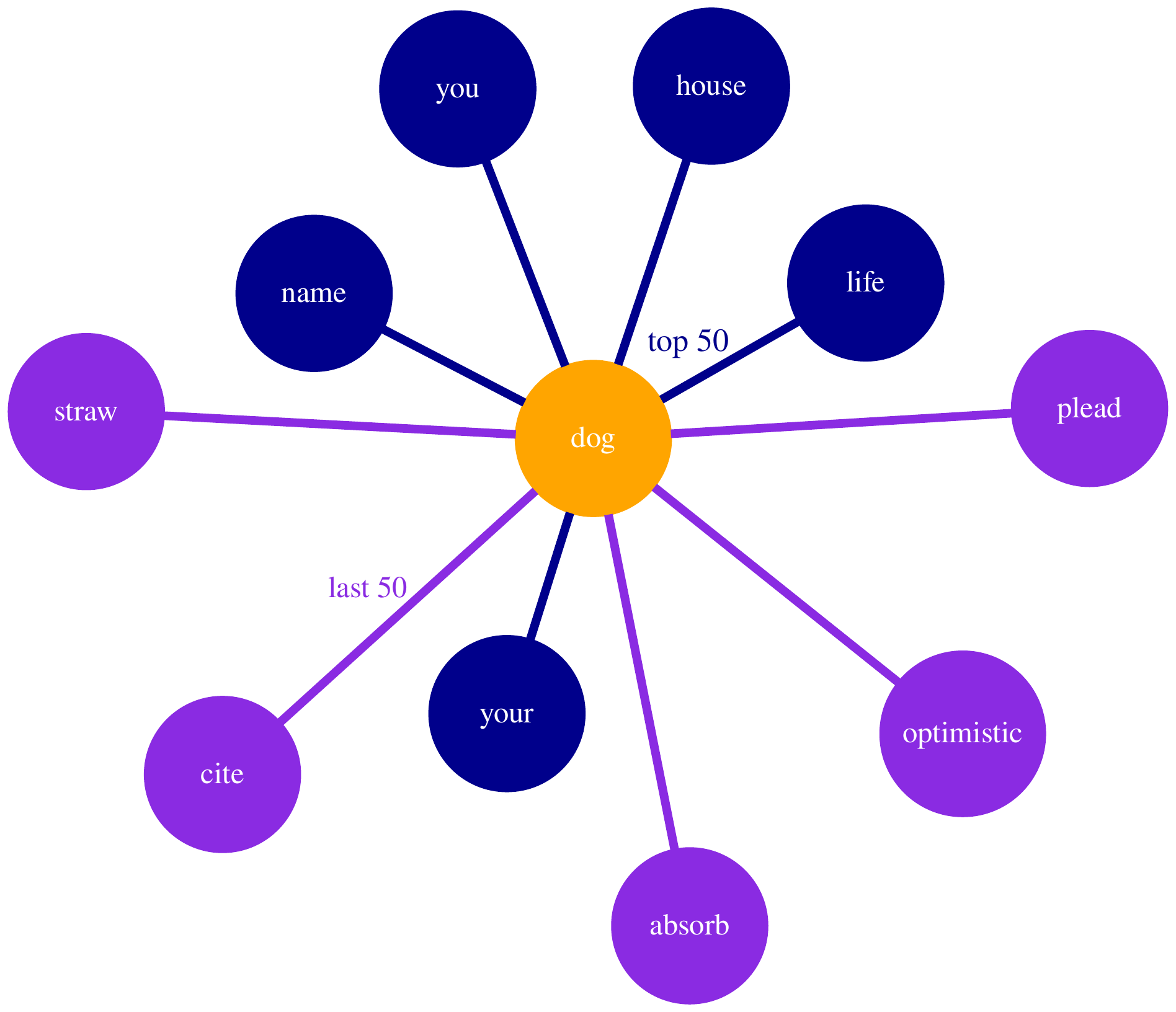}
        \caption{}
        \label{fig:ex1}
    \end{subfigure}%
    \hfill%
    \begin{subfigure}[t]{0.33\linewidth}
        \centering
        \includegraphics[width=\linewidth]{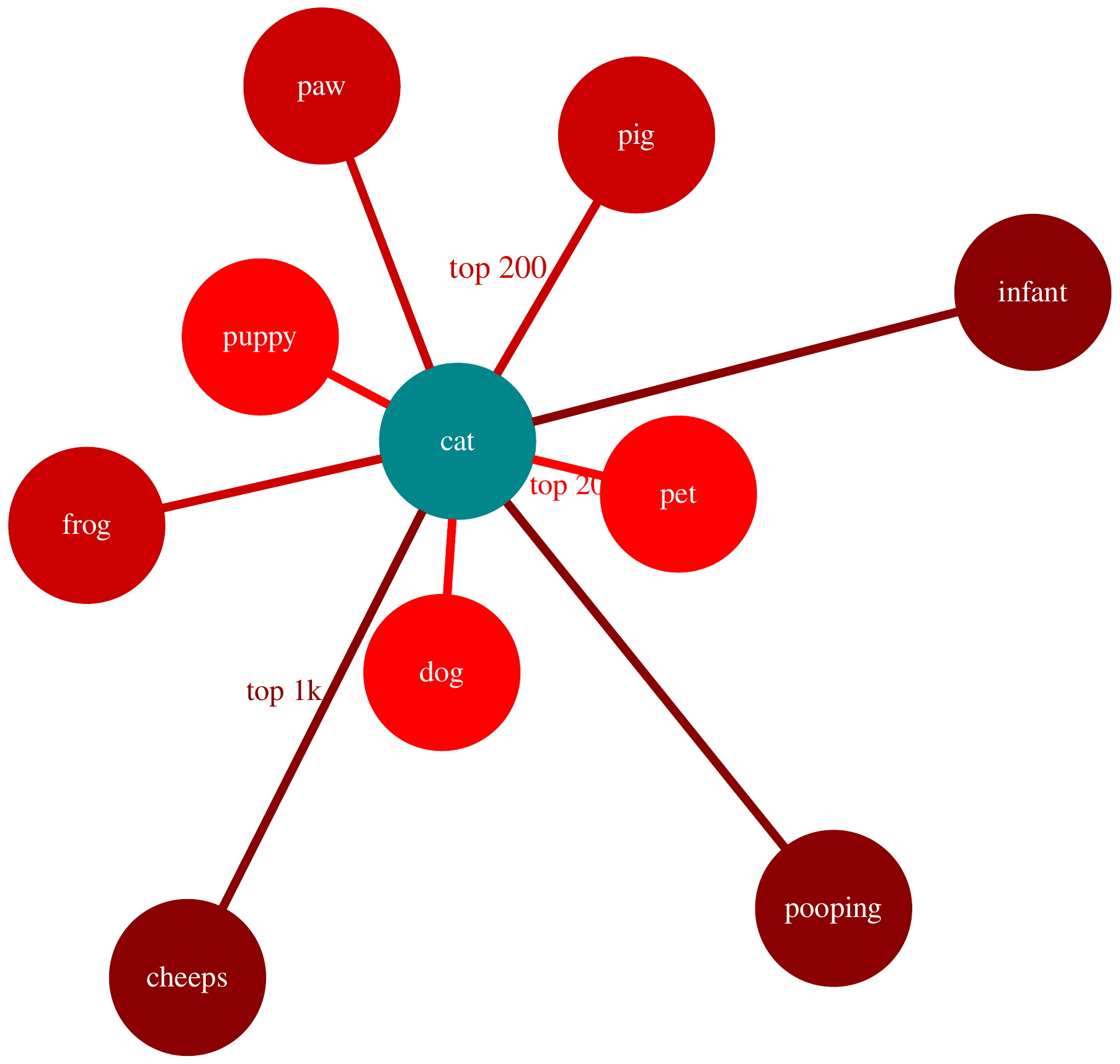}
        \caption{}
        \label{fig:ex3}
    \end{subfigure}%
    \hfill%
    \begin{subfigure}[t]{0.33\linewidth}
        \centering
        \includegraphics[width=\linewidth]{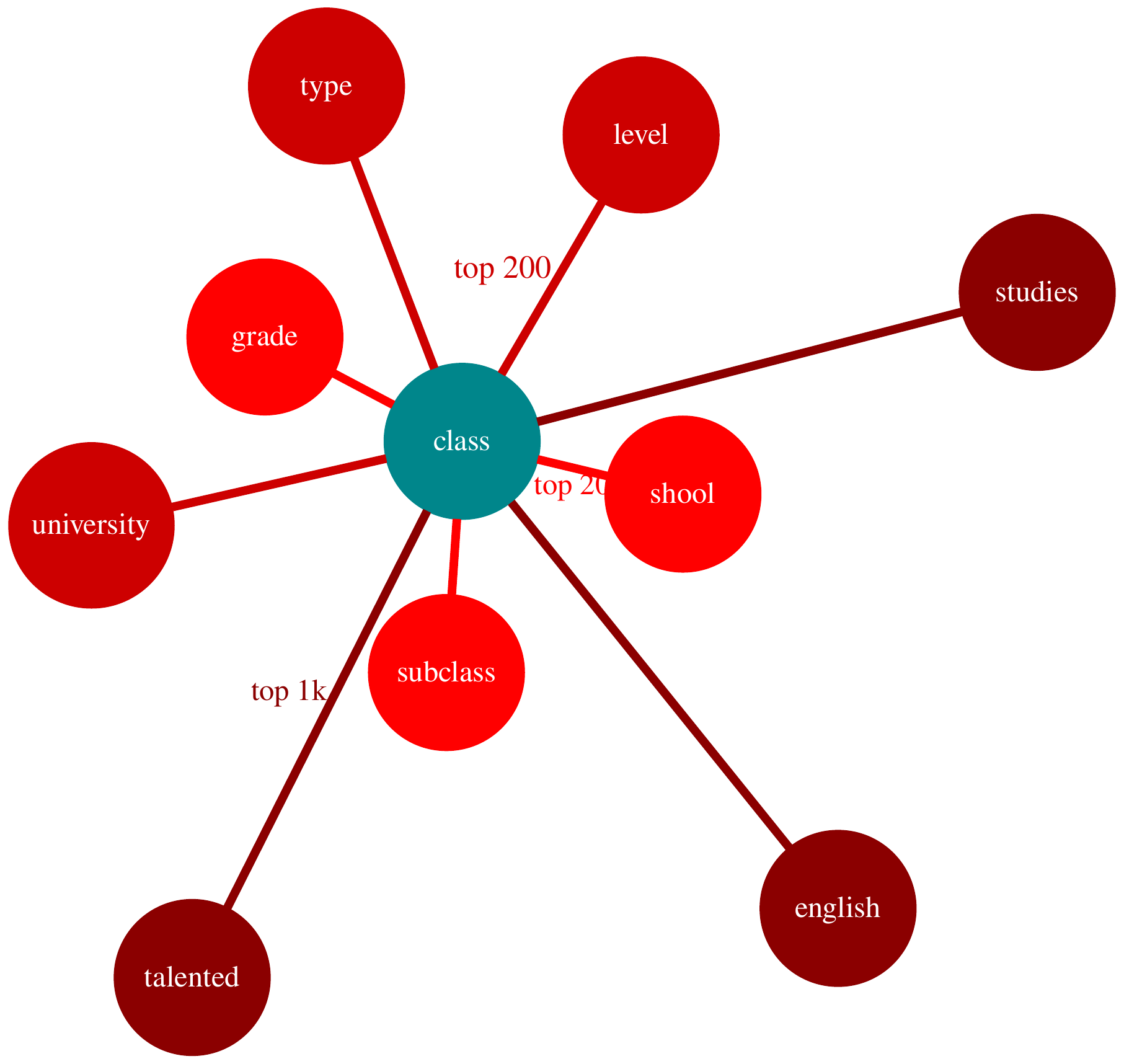}
        \caption{}
        \label{fig:ex2}
    \end{subfigure}%
    \caption{\textbf{(a) Concepts that have high and low associations to concept \emph{dog}; (b) Concepts that have high similarity to concept \emph{cat}; (c) Concepts that have high similarity to concept \emph{class}, with the potential for the from the semantic subspace of \emph{education-related-concepts} (\eg, \emph{school} and \emph{grade}) to \emph{computer-related-concepts} (\eg, \emph{subclass} and \emph{type}).}}
\end{figure}

\paragraph{Acknowledgement}

The authors thank Mr. David Turner for inspiring and helpful discussions on designing Mindle.

{
    \small
    \bibliographystyle{apalike}
    \bibliography{references}
}

\clearpage
\appendix

\section{User interface of Mindle}

\begin{figure}[ht]
\begin{subfigure}{.6\textwidth}
  \centering
  \includegraphics[width=.9\linewidth]{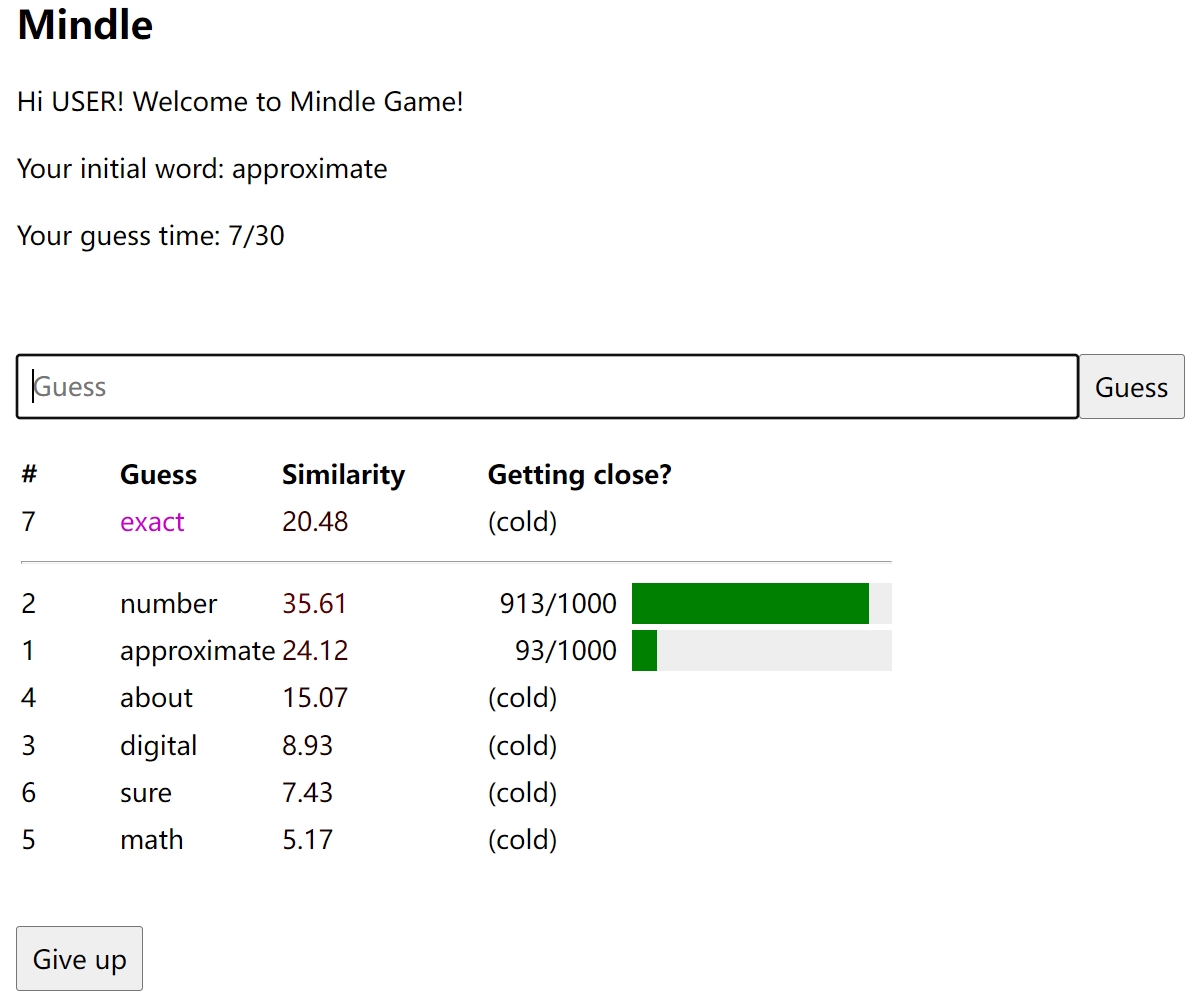}  
  \caption{}
  \label{fig:mindle-web}
\end{subfigure}
\begin{subfigure}{.4\textwidth}
  \centering
  \includegraphics[width=.7\linewidth, height=1\linewidth]{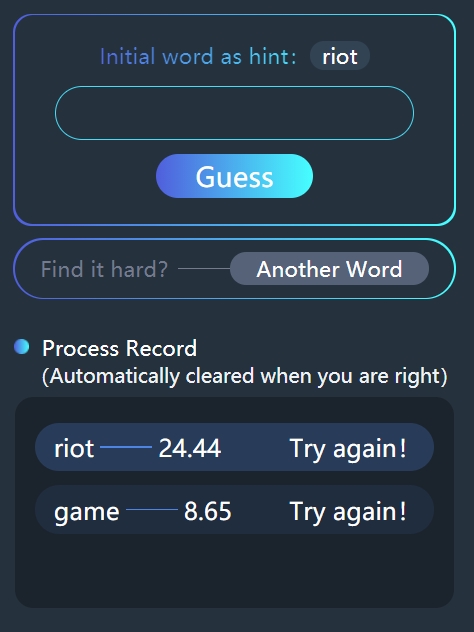}  
  \caption{}
  \label{fig:mindle-mobile}
\end{subfigure}
\caption{\textbf{User interface of Mindle.} (a) Web-based Mindle; (b) Mobile-based Mindle}
\end{figure}

Given a starting word, the users are expected to navigate toward a secret target word. Users travel in the semantic world by guessing words. For each time jumping to a word, the users will get a similarity score indicating the distance between their current position and the target. For the sake of Mindle's pilot study, we designed two versions of our Mindle game. Especially, the Web-based Mindle is designed for both lab-based and online experiments (see \cref{fig:mindle-web}), and the Mobile-based Mindle can be accessed from mobile terminals, making it suitable for larger-scale online experiments (see \cref{fig:mindle-mobile}).

\section{Examples of conceptual knowledge representation}

\cref{tab:cn-table} shows some canonical examples of conceptual knowledge representation. Each example shows the highly associated concepts to the centered concept. A set of stop words have been removed to make the results more representative.

\begin{table}[!h]
\centering
\caption{\textbf{Examples of top-associated conceptual knowledge representation}}
\begin{tabular}{c|c} \hline
Word & Top associated concept\\\hline\hline
school               & building, county, education,
  national, community, family, government, city, house, university  \\\hline
cat                  & life, name, family, medal, known, me, house, into, your, time                                               \\\hline
work                 & press,have,government,community, title, zone, action,
  field, works, working, more, time     \\\hline
apple                & park, famous, story, award, album, music, different, out, title                                             \\\hline
train                & tour, population, district, project, zone,~                                                                 \\\hline
gift                 & what, new, different, 'something, house, special, story, inside                                             \\    \hline                                                                           
\end{tabular}\label{tab:cn-table}
\end{table}
\definecolor{top1}{rgb}{1,0.439,0.439}
\definecolor{top10}{rgb}{0.996,0.663,0.663} 
\definecolor{top100}{rgb}{0.992,0.788,0.788} 
\definecolor{top500}{rgb}{1,0.898,0.898} 
\section{Examples of score contour line}
\begin{table}
\caption{\textbf{The score contour line based on the cos similarity.}\\
\colorbox{top1}{Top 1} \colorbox{top10}{Top 10} \colorbox{top100}{Top 100} \colorbox{top500}{Top 500}}
\begin{subtable}{0.5\linewidth}
\centering
\caption{\textbf{cat}}
\arrayrulecolor{black}
\begin{tabular}[t]{c|c|c} 
\hline
\rowcolor[rgb]{0.812,0.808,0.808} Target                                                 & Word          & Cos Similarity  \\ 
\hline
\rowcolor[rgb]{1,0.439,0.439} {\cellcolor[rgb]{0.808,0.808,0.808}}                       & cat           & 1           \\ 
\hhline{|>{\arrayrulecolor[rgb]{0.808,0.808,0.808}}->{\arrayrulecolor{black}}--|}
\rowcolor[rgb]{0.996,0.663,0.663} {\cellcolor[rgb]{0.808,0.808,0.808}}                   & cats          & 0.809938    \\ 
\hhline{|>{\arrayrulecolor[rgb]{0.808,0.808,0.808}}->{\arrayrulecolor{black}}--|}
\rowcolor[rgb]{0.996,0.663,0.663} {\cellcolor[rgb]{0.808,0.808,0.808}}                   & dog           & 0.760946    \\ 
\hhline{|>{\arrayrulecolor[rgb]{0.808,0.808,0.808}}->{\arrayrulecolor{black}}--|}
\rowcolor[rgb]{0.996,0.663,0.663} {\cellcolor[rgb]{0.808,0.808,0.808}}                   & kitten        & 0.746498    \\ 
\hhline{|>{\arrayrulecolor[rgb]{0.808,0.808,0.808}}->{\arrayrulecolor{black}}--|}
\rowcolor[rgb]{0.996,0.663,0.663} {\cellcolor[rgb]{0.808,0.808,0.808}}                   & feline        & 0.732624    \\ 
\hhline{|>{\arrayrulecolor[rgb]{0.808,0.808,0.808}}->{\arrayrulecolor{black}}--|}
\rowcolor[rgb]{0.996,0.663,0.663} {\cellcolor[rgb]{0.808,0.808,0.808}}                   & beagle        & 0.715058    \\ 
\hhline{|>{\arrayrulecolor[rgb]{0.808,0.808,0.808}}->{\arrayrulecolor{black}}--|}
\rowcolor[rgb]{0.996,0.663,0.663} {\cellcolor[rgb]{0.808,0.808,0.808}}                   & puppy         & 0.707545    \\ 
\hhline{|>{\arrayrulecolor[rgb]{0.808,0.808,0.808}}->{\arrayrulecolor{black}}--|}
\rowcolor[rgb]{0.996,0.663,0.663} {\cellcolor[rgb]{0.808,0.808,0.808}}                   & pup           & 0.693429    \\ 
\hhline{|>{\arrayrulecolor[rgb]{0.808,0.808,0.808}}->{\arrayrulecolor{black}}--|}
\rowcolor[rgb]{0.996,0.663,0.663} {\cellcolor[rgb]{0.808,0.808,0.808}}                   & pet           & 0.689153    \\ 
\hhline{|>{\arrayrulecolor[rgb]{0.808,0.808,0.808}}->{\arrayrulecolor{black}}--|}
\rowcolor[rgb]{0.996,0.663,0.663} {\cellcolor[rgb]{0.808,0.808,0.808}}                   & felines       & 0.675593    \\ 
\hhline{|>{\arrayrulecolor[rgb]{0.808,0.808,0.808}}->{\arrayrulecolor{black}}--|}
\rowcolor[rgb]{0.996,0.663,0.663} {\cellcolor[rgb]{0.808,0.808,0.808}}                   & chihuahua     & 0.670976    \\ 
\hhline{|>{\arrayrulecolor[rgb]{0.808,0.808,0.808}}->{\arrayrulecolor{black}}--|}
\rowcolor[rgb]{0.992,0.788,0.788} {\cellcolor[rgb]{0.808,0.808,0.808}}                   & bassets       & 0.520498    \\ 
\hhline{|>{\arrayrulecolor[rgb]{0.808,0.808,0.808}}->{\arrayrulecolor{black}}--|}
\rowcolor[rgb]{0.992,0.788,0.788} {\cellcolor[rgb]{0.808,0.808,0.808}}                   & rooster       & 0.51893     \\ 
\hhline{|>{\arrayrulecolor[rgb]{0.808,0.808,0.808}}->{\arrayrulecolor{black}}--|}
\rowcolor[rgb]{0.992,0.788,0.788} {\cellcolor[rgb]{0.808,0.808,0.808}}                   & owl           & 0.518349    \\ 
\hhline{|>{\arrayrulecolor[rgb]{0.808,0.808,0.808}}->{\arrayrulecolor{black}}--|}
\rowcolor[rgb]{0.992,0.788,0.788} {\cellcolor[rgb]{0.808,0.808,0.808}}                   & pinscher      & 0.517772    \\ 
\hhline{|>{\arrayrulecolor[rgb]{0.808,0.808,0.808}}->{\arrayrulecolor{black}}--|}
\rowcolor[rgb]{0.992,0.788,0.788} {\cellcolor[rgb]{0.808,0.808,0.808}}                   & tiger         & 0.517296    \\ 
\hhline{|>{\arrayrulecolor[rgb]{0.808,0.808,0.808}}->{\arrayrulecolor{black}}--|}
\rowcolor[rgb]{0.992,0.788,0.788} {\cellcolor[rgb]{0.808,0.808,0.808}}                   & piglet        & 0.516684    \\ 
\hhline{|>{\arrayrulecolor[rgb]{0.808,0.808,0.808}}->{\arrayrulecolor{black}}--|}
\rowcolor[rgb]{0.992,0.788,0.788} {\cellcolor[rgb]{0.808,0.808,0.808}}                   & kelpie        & 0.515803    \\ 
\hhline{|>{\arrayrulecolor[rgb]{0.808,0.808,0.808}}->{\arrayrulecolor{black}}--|}
\rowcolor[rgb]{0.992,0.788,0.788} {\cellcolor[rgb]{0.808,0.808,0.808}}                   & dachshunds    & 0.515763    \\ 
\hhline{|>{\arrayrulecolor[rgb]{0.808,0.808,0.808}}->{\arrayrulecolor{black}}--|}
\rowcolor[rgb]{0.992,0.788,0.788} {\cellcolor[rgb]{0.808,0.808,0.808}}                   & schnauzers    & 0.514645    \\ 
\hhline{|>{\arrayrulecolor[rgb]{0.808,0.808,0.808}}->{\arrayrulecolor{black}}--|}
\rowcolor[rgb]{0.992,0.788,0.788} {\cellcolor[rgb]{0.808,0.808,0.808}}                   & bird          & 0.514626    \\ 
\hhline{|>{\arrayrulecolor[rgb]{0.808,0.808,0.808}}->{\arrayrulecolor{black}}--|}
\rowcolor[rgb]{1,0.898,0.898} {\cellcolor[rgb]{0.808,0.808,0.808}}                       & earless       & 0.392231    \\ 
\hhline{|>{\arrayrulecolor[rgb]{0.808,0.808,0.808}}->{\arrayrulecolor{black}}--|}
\rowcolor[rgb]{1,0.898,0.898} {\cellcolor[rgb]{0.808,0.808,0.808}}                       & hoarder       & 0.392067    \\ 
\hhline{|>{\arrayrulecolor[rgb]{0.808,0.808,0.808}}->{\arrayrulecolor{black}}--|}
\rowcolor[rgb]{1,0.898,0.898} {\cellcolor[rgb]{0.808,0.808,0.808}}                       & lynx          & 0.392044    \\ 
\hhline{|>{\arrayrulecolor[rgb]{0.808,0.808,0.808}}->{\arrayrulecolor{black}}--|}
\rowcolor[rgb]{1,0.898,0.898} {\cellcolor[rgb]{0.808,0.808,0.808}}                       & shrike        & 0.392036    \\ 
\hhline{|>{\arrayrulecolor[rgb]{0.808,0.808,0.808}}->{\arrayrulecolor{black}}--|}
\rowcolor[rgb]{1,0.898,0.898} {\cellcolor[rgb]{0.808,0.808,0.808}}                       & panleukopenia & 0.391696    \\ 
\hhline{|>{\arrayrulecolor[rgb]{0.808,0.808,0.808}}->{\arrayrulecolor{black}}--|}
\rowcolor[rgb]{1,0.898,0.898} {\cellcolor[rgb]{0.808,0.808,0.808}}                       & iguanas       & 0.391474    \\ 
\hhline{|>{\arrayrulecolor[rgb]{0.808,0.808,0.808}}->{\arrayrulecolor{black}}--|}
\rowcolor[rgb]{1,0.898,0.898} {\cellcolor[rgb]{0.808,0.808,0.808}}                       & doglike       & 0.391446    \\ 
\hhline{|>{\arrayrulecolor[rgb]{0.808,0.808,0.808}}->{\arrayrulecolor{black}}--|}
\rowcolor[rgb]{1,0.898,0.898} {\cellcolor[rgb]{0.808,0.808,0.808}}                       & yelping       & 0.391318    \\ 
\hhline{|>{\arrayrulecolor[rgb]{0.808,0.808,0.808}}->{\arrayrulecolor{black}}--|}
\rowcolor[rgb]{1,0.898,0.898} {\cellcolor[rgb]{0.808,0.808,0.808}}                       & crow          & 0.391283    \\ 
\hhline{|>{\arrayrulecolor[rgb]{0.808,0.808,0.808}}->{\arrayrulecolor{black}}--|}
\rowcolor[rgb]{1,0.898,0.898} \multirow{-31}{*}{{\cellcolor[rgb]{0.808,0.808,0.808}}cat} & rabbity       & 0.391252    \\
\hline
\end{tabular}
\end{subtable}%
\begin{subtable}{0.5\linewidth}
\centering
\caption{\textbf{green}}
\arrayrulecolor{black}
\begin{tabular}[t]{c|c|c} 
\hline
\rowcolor[rgb]{0.812,0.808,0.808} Target                                                   & Word         & Cos Similarity  \\ 
\hline
\rowcolor[rgb]{1,0.439,0.439} {\cellcolor[rgb]{0.808,0.808,0.808}}                         & green        & 1           \\ 
\hhline{|>{\arrayrulecolor[rgb]{0.808,0.808,0.808}}->{\arrayrulecolor{black}}--|}
\rowcolor[rgb]{0.996,0.663,0.663} {\cellcolor[rgb]{0.808,0.808,0.808}}                     & greener      & 0.809938    \\ 
\hhline{|>{\arrayrulecolor[rgb]{0.808,0.808,0.808}}->{\arrayrulecolor{black}}--|}
\rowcolor[rgb]{0.996,0.663,0.663} {\cellcolor[rgb]{0.808,0.808,0.808}}                     & red          & 0.760946    \\ 
\hhline{|>{\arrayrulecolor[rgb]{0.808,0.808,0.808}}->{\arrayrulecolor{black}}--|}
\rowcolor[rgb]{0.996,0.663,0.663} {\cellcolor[rgb]{0.808,0.808,0.808}}                     & greening     & 0.746498    \\ 
\hhline{|>{\arrayrulecolor[rgb]{0.808,0.808,0.808}}->{\arrayrulecolor{black}}--|}
\rowcolor[rgb]{0.996,0.663,0.663} {\cellcolor[rgb]{0.808,0.808,0.808}}                     & yellow       & 0.732624    \\ 
\hhline{|>{\arrayrulecolor[rgb]{0.808,0.808,0.808}}->{\arrayrulecolor{black}}--|}
\rowcolor[rgb]{0.996,0.663,0.663} {\cellcolor[rgb]{0.808,0.808,0.808}}                     & blue         & 0.715058    \\ 
\hhline{|>{\arrayrulecolor[rgb]{0.808,0.808,0.808}}->{\arrayrulecolor{black}}--|}
\rowcolor[rgb]{0.996,0.663,0.663} {\cellcolor[rgb]{0.808,0.808,0.808}}                     & brown        & 0.707545    \\ 
\hhline{|>{\arrayrulecolor[rgb]{0.808,0.808,0.808}}->{\arrayrulecolor{black}}--|}
\rowcolor[rgb]{0.996,0.663,0.663} {\cellcolor[rgb]{0.808,0.808,0.808}}                     & florescent   & 0.693429    \\ 
\hhline{|>{\arrayrulecolor[rgb]{0.808,0.808,0.808}}->{\arrayrulecolor{black}}--|}
\rowcolor[rgb]{0.996,0.663,0.663} {\cellcolor[rgb]{0.808,0.808,0.808}}                     & greenest     & 0.689153    \\ 
\hhline{|>{\arrayrulecolor[rgb]{0.808,0.808,0.808}}->{\arrayrulecolor{black}}--|}
\rowcolor[rgb]{0.996,0.663,0.663} {\cellcolor[rgb]{0.808,0.808,0.808}}                     & nongreen     & 0.675593    \\ 
\hhline{|>{\arrayrulecolor[rgb]{0.808,0.808,0.808}}->{\arrayrulecolor{black}}--|}
\rowcolor[rgb]{0.996,0.663,0.663} {\cellcolor[rgb]{0.808,0.808,0.808}}                     & purple       & 0.670976    \\ 
\hhline{|>{\arrayrulecolor[rgb]{0.808,0.808,0.808}}->{\arrayrulecolor{black}}--|}
\rowcolor[rgb]{0.992,0.788,0.788} {\cellcolor[rgb]{0.808,0.808,0.808}}                     & pistache     & 0.520498    \\ 
\hhline{|>{\arrayrulecolor[rgb]{0.808,0.808,0.808}}->{\arrayrulecolor{black}}--|}
\rowcolor[rgb]{0.992,0.788,0.788} {\cellcolor[rgb]{0.808,0.808,0.808}}                     & echeveria    & 0.51893     \\ 
\hhline{|>{\arrayrulecolor[rgb]{0.808,0.808,0.808}}->{\arrayrulecolor{black}}--|}
\rowcolor[rgb]{0.992,0.788,0.788} {\cellcolor[rgb]{0.808,0.808,0.808}}                     & paspalum     & 0.518349    \\ 
\hhline{|>{\arrayrulecolor[rgb]{0.808,0.808,0.808}}->{\arrayrulecolor{black}}--|}
\rowcolor[rgb]{0.992,0.788,0.788} {\cellcolor[rgb]{0.808,0.808,0.808}}                     & greened      & 0.517772    \\ 
\hhline{|>{\arrayrulecolor[rgb]{0.808,0.808,0.808}}->{\arrayrulecolor{black}}--|}
\rowcolor[rgb]{0.992,0.788,0.788} {\cellcolor[rgb]{0.808,0.808,0.808}}                     & bicolored    & 0.517296    \\ 
\hhline{|>{\arrayrulecolor[rgb]{0.808,0.808,0.808}}->{\arrayrulecolor{black}}--|}
\rowcolor[rgb]{0.992,0.788,0.788} {\cellcolor[rgb]{0.808,0.808,0.808}}                     & multicolored & 0.516684    \\ 
\hhline{|>{\arrayrulecolor[rgb]{0.808,0.808,0.808}}->{\arrayrulecolor{black}}--|}
\rowcolor[rgb]{0.992,0.788,0.788} {\cellcolor[rgb]{0.808,0.808,0.808}}                     & brittlebush  & 0.515803    \\ 
\hhline{|>{\arrayrulecolor[rgb]{0.808,0.808,0.808}}->{\arrayrulecolor{black}}--|}
\rowcolor[rgb]{0.992,0.788,0.788} {\cellcolor[rgb]{0.808,0.808,0.808}}                     & arborvitaes  & 0.515763    \\ 
\hhline{|>{\arrayrulecolor[rgb]{0.808,0.808,0.808}}->{\arrayrulecolor{black}}--|}
\rowcolor[rgb]{0.992,0.788,0.788} {\cellcolor[rgb]{0.808,0.808,0.808}}                     & stripes      & 0.514645    \\ 
\hhline{|>{\arrayrulecolor[rgb]{0.808,0.808,0.808}}->{\arrayrulecolor{black}}--|}
\rowcolor[rgb]{0.992,0.788,0.788} {\cellcolor[rgb]{0.808,0.808,0.808}}                     & sienna       & 0.514626    \\ 
\hhline{|>{\arrayrulecolor[rgb]{0.808,0.808,0.808}}->{\arrayrulecolor{black}}--|}
\rowcolor[rgb]{1,0.898,0.898} {\cellcolor[rgb]{0.808,0.808,0.808}}                         & conserve     & 0.392231    \\ 
\hhline{|>{\arrayrulecolor[rgb]{0.808,0.808,0.808}}->{\arrayrulecolor{black}}--|}
\rowcolor[rgb]{1,0.898,0.898} {\cellcolor[rgb]{0.808,0.808,0.808}}                         & photovoltaic & 0.392067    \\ 
\hhline{|>{\arrayrulecolor[rgb]{0.808,0.808,0.808}}->{\arrayrulecolor{black}}--|}
\rowcolor[rgb]{1,0.898,0.898} {\cellcolor[rgb]{0.808,0.808,0.808}}                         & leafier      & 0.392044    \\ 
\hhline{|>{\arrayrulecolor[rgb]{0.808,0.808,0.808}}->{\arrayrulecolor{black}}--|}
\rowcolor[rgb]{1,0.898,0.898} {\cellcolor[rgb]{0.808,0.808,0.808}}                         & euonymous    & 0.392036    \\ 
\hhline{|>{\arrayrulecolor[rgb]{0.808,0.808,0.808}}->{\arrayrulecolor{black}}--|}
\rowcolor[rgb]{1,0.898,0.898} {\cellcolor[rgb]{0.808,0.808,0.808}}                         & alpenglow    & 0.391696    \\ 
\hhline{|>{\arrayrulecolor[rgb]{0.808,0.808,0.808}}->{\arrayrulecolor{black}}--|}
\rowcolor[rgb]{1,0.898,0.898} {\cellcolor[rgb]{0.808,0.808,0.808}}                         & coppery      & 0.391474    \\ 
\hhline{|>{\arrayrulecolor[rgb]{0.808,0.808,0.808}}->{\arrayrulecolor{black}}--|}
\rowcolor[rgb]{1,0.898,0.898} {\cellcolor[rgb]{0.808,0.808,0.808}}                         & tomatillo    & 0.391446    \\ 
\hhline{|>{\arrayrulecolor[rgb]{0.808,0.808,0.808}}->{\arrayrulecolor{black}}--|}
\rowcolor[rgb]{1,0.898,0.898} {\cellcolor[rgb]{0.808,0.808,0.808}}                         & beautifying  & 0.391318    \\ 
\hhline{|>{\arrayrulecolor[rgb]{0.808,0.808,0.808}}->{\arrayrulecolor{black}}--|}
\rowcolor[rgb]{1,0.898,0.898} {\cellcolor[rgb]{0.808,0.808,0.808}}                         & marram       & 0.391283    \\ 
\hhline{|>{\arrayrulecolor[rgb]{0.808,0.808,0.808}}->{\arrayrulecolor{black}}--|}
\rowcolor[rgb]{1,0.898,0.898} \multirow{-31}{*}{{\cellcolor[rgb]{0.808,0.808,0.808}}green} & tangelo      & 0.391252    \\
\hline
\end{tabular}
\end{subtable}
\label{tab:similarity}
\end{table}
\cref{tab:similarity} shows some canonical examples of the similarity-based semantic landscape. A contour line is shaped by different concepts with the same-level score to the target concept. Each example shows a series of contour lines by different scores. Concept \emph{cat} has a single major semantic meaning, while \emph{green} has two major semantic meanings. Hence, the similar concepts to \emph{cat} lie in a single semantic subspace, while those to green lie in two semantic subspaces (color and plant).

\section{Player trajectories}
\begin{figure}[!ht]
\begin{subfigure}{\textwidth}
  \centering
  \includegraphics[width=0.9\linewidth]{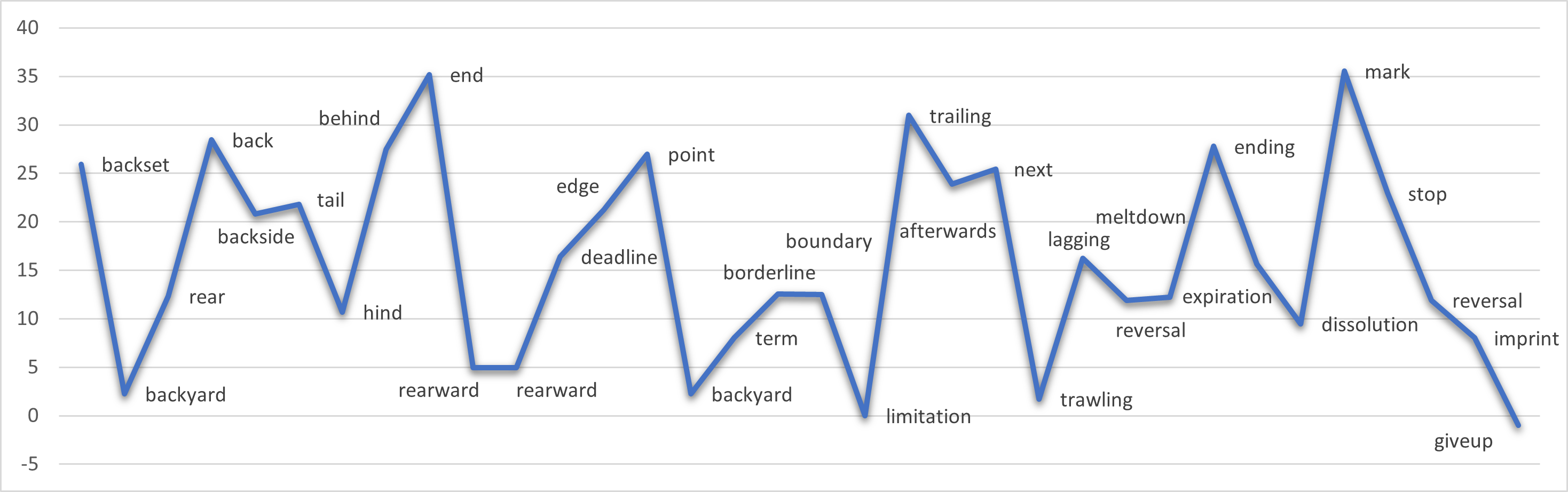} 
  \caption{Target Word: \textit{Finish}}
  \label{fig:finish}
\end{subfigure}\\
\begin{subfigure}{\textwidth}
  \centering
  \includegraphics[width=0.9\linewidth]{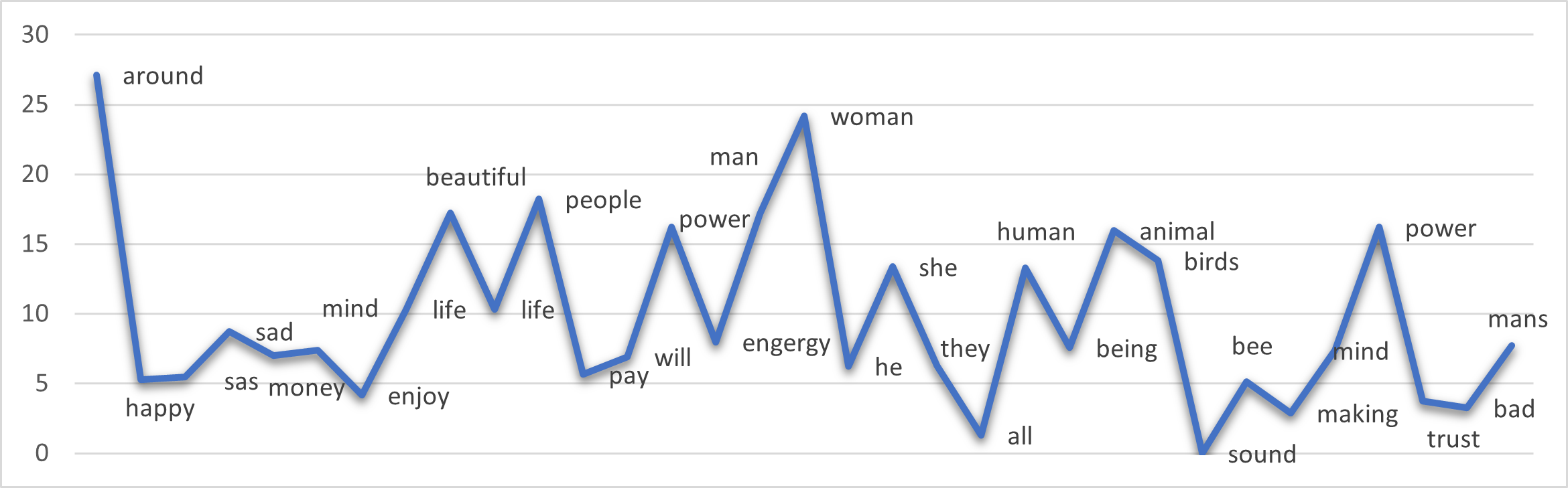} 
  \caption{Target Word: \textit{Northest}}
  \label{fig:northest}
\end{subfigure}\\
\begin{subfigure}{\textwidth}
  \centering
  \includegraphics[width=0.9\linewidth]{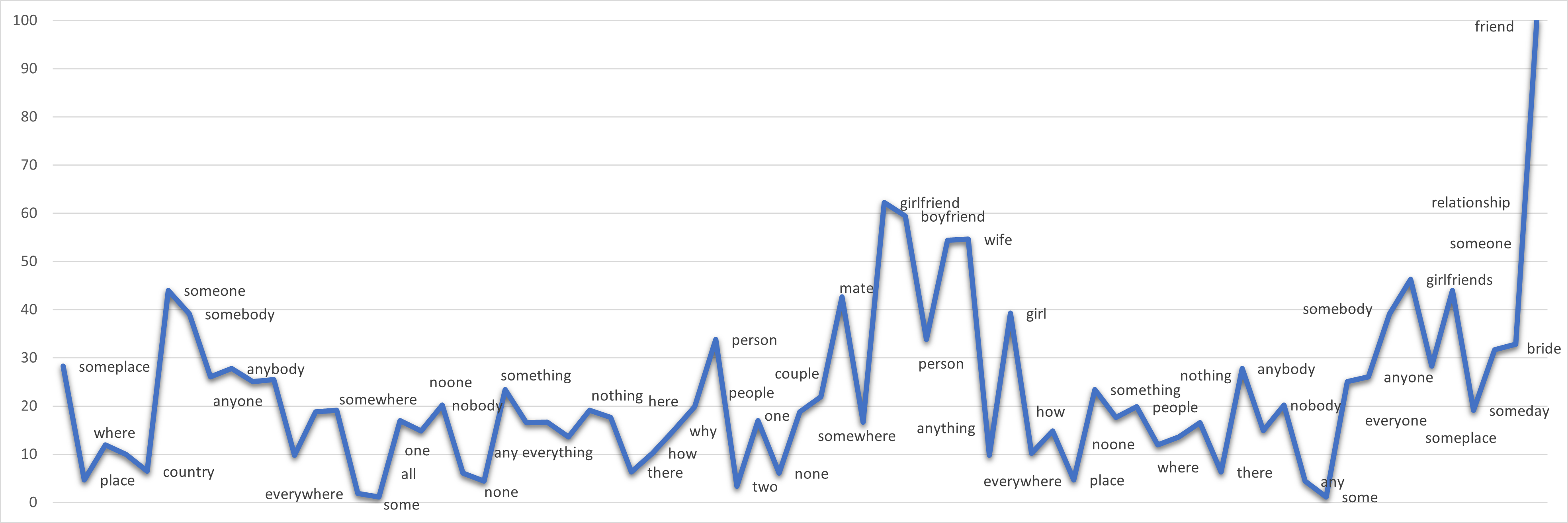} 
  \caption{Target Word: \textit{Friend}}
  \label{fig:friend}
\end{subfigure}
\caption{\textbf{Examples of player trajectories in Mindle.}}
\label{fig:trajectory}
\end{figure}

\Cref{fig:trajectory} shows some exemplar trajectories generated by players. Thinking patterns mentioned in the paper, such as \emph{insight-seeking} and \emph{domain-knowledge-relying} can be clearly observed in these trajectories. Besides, several \aka \emph{Aha! moment} can be observed in the test process.

\end{document}